\documentclass[journal,twoside, web]{IEEEtran}
\usepackage{tmi}
\usepackage{cite}
\usepackage{amsmath,amssymb,amsfonts}
\usepackage{algorithmic}
\usepackage{graphicx}
\usepackage{textcomp}
\usepackage{hyperref}       
\usepackage{url}            
\usepackage{booktabs}       
\usepackage{nicefrac}       
\usepackage{microtype}      
\usepackage{epstopdf}
\DeclareMathOperator{\E}{\mathbb{E}}

\usepackage{etoolbox}
\makeatletter
\patchcmd{\@makecaption}
  {\scshape}
  {}
  {}
  {}
\makeatother

\def\BibTeX{{\rm B\kern-.05em{\sc i\kern-.025em b}\kern-.08em
    T\kern-.1667em\lower.7ex\hbox{E}\kern-.125emX}}
\begin{document}
\title{ICAM-reg: Interpretable Classification and Regression with Feature Attribution for Mapping Neurological Phenotypes in Individual Scans}
\author{Cher Bass, Mariana da Silva, Carole Sudre, Logan Z. J. Williams, Petru-Daniel Tudosiu, Fidel Alfaro-Almagro, Sean P. Fitzgibbon, Matthew F. Glasser, Stephen M. Smith, Emma C. Robinson, and the Alzheimer’s Disease Neuroimaging Initiative
\thanks{The work of E.C.R. and C.B. was supported by the Academy of Medical Sciences/the British Heart Foundation/the Government Department of Business, Energy and Industrial Strategy/the Wellcome Trust Springboard Award [SBF003/1116] and E.C.R., C.B. and S.M.S. are supported by Wellcome Collaborative Award  [215573/Z/19/Z]. PD.T. was supported by the EPSRC Research Council, part of the EPSRC DTP [EP/R513064/1]. M. DS. would like to acknowledge funding from the EPSRC Centre for Doctoral Training in Smart Medical Imaging [EP/S022104/1]. L.Z.J.W would like to acknowledge funding from the Commonwealth Scholarship Commission, United Kingdom.}
\thanks{The UK Biobank data was accessed under Application Number 8107.}
\thanks{ADNI data used in preparation of this article were obtained from the Alzheimer’s Disease Neuroimaging Initiative (ADNI) database (adni.loni.usc.edu). As such, the investigators within the ADNI contributed to the design and implementation of ADNI and/or provided data but did not participate in analysis or writing of this report. A complete listing of ADNI investigators can be found at: \url{http://adni.loni.usc.edu/wp-content/uploads/how_to_apply/ADNI_Acknowledgement_List.pdf}. The ADNI data used in this work was funded by the Alzheimer's Disease Neuroimaging Initiative (ADNI) (National Institutes of Health Grant U01 AG024904) and DOD ADNI (Department of Defense award number W81XWH-12-2-0012).}
}

\maketitle

\begin{abstract}
An important goal of medical imaging is to be able to precisely detect patterns of disease specific to individual scans; however, this is challenged in brain imaging by the degree of heterogeneity of shape and appearance. Traditional methods, based on image registration to a global template, historically fail to detect variable features of disease, as they utilise population-based analyses, suited primarily to studying group-average effects. In this paper we therefore take advantage of recent developments in generative deep learning to develop a method for simultaneous classification, or regression, and feature attribution (FA). Specifically, we explore the use of a VAE-GAN translation network called ICAM, to explicitly disentangle class relevant features from background confounds for improved interpretability and regression of neurological phenotypes. We validate our method on the tasks of Mini-Mental State Examination (MMSE) cognitive test score prediction for the Alzheimer’s Disease Neuroimaging Initiative (ADNI) cohort, as well as brain age prediction, for both neurodevelopment and neurodegeneration, using the developing Human Connectome Project (dHCP) and UK Biobank datasets. We show that the generated FA maps can be used to explain outlier predictions and demonstrate that the inclusion of a regression module improves the disentanglement of the latent space. Our code is freely available on Github \url{https://github.com/CherBass/ICAM}.

\end{abstract}


\section{Introduction}
\label{sec:introduction}

\begin{figure*}[!htb]
  \centering
\makebox[\linewidth]{
	\includegraphics[width=0.8\textwidth]{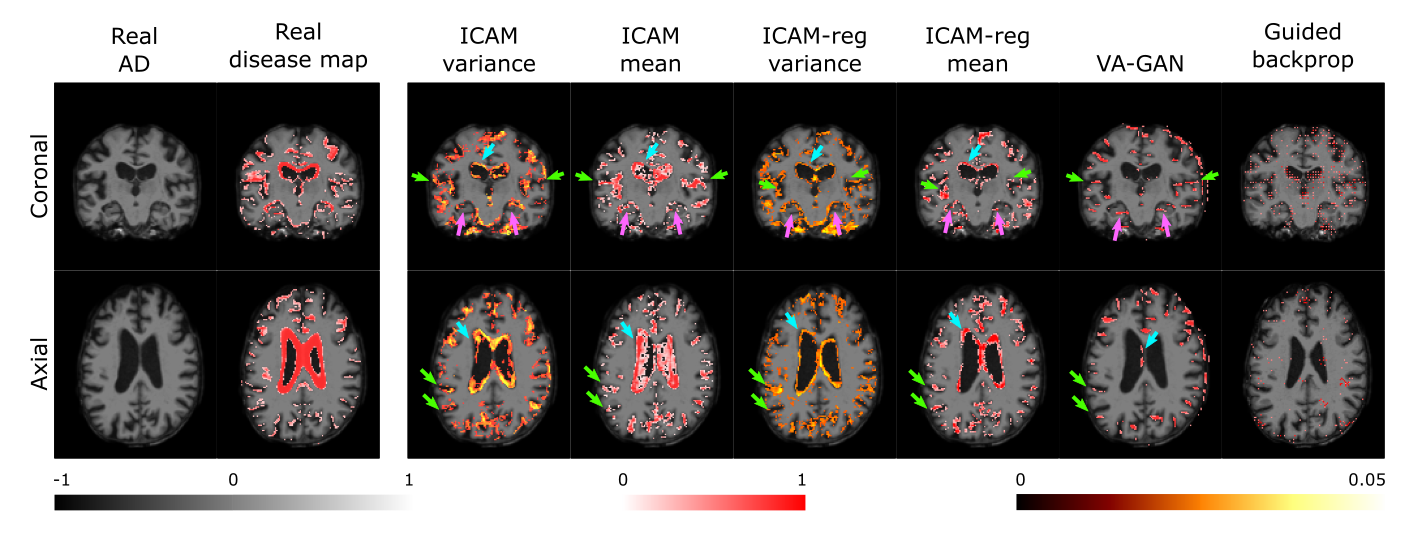}}
\caption{ADNI comparisons of Feature Attribution (FA) maps. ICAM is the first known method able to generate variance and mean FA maps in test time, and shows good detection of the ventricles (blue arrows), cortex (green arrows), and hippocampus (pink arrows) when compared with the ground truth disease map. The top 2 baseline methods are shown here, and perform sub-optimally in comparison to ICAM.}
\label{fig:adni_comparison}
\end{figure*}

Brain images represent a significant resource in the development of mechanistic models of behaviour and neurological/psychiatric disease as, in principle, they capture measurable neuroanatomical traits that are heritable, present in unaffected siblings and detectable prior to disease onset \cite{cullen2019polygenic}. For many complex disorders, however, these features of disease  \cite{iqbal2005subgroups,ross2006neurobiology} are subtle, variable and obscured by a back-drop of significant natural variation in brain shape and appearance \cite{glasser2016multi,kong2019spatial}; this makes them extremely difficult to detect. 

Traditional approaches for analysis of brain magnetic resonance imaging (MRI) rely on group-wise comparisons between disease and control groups, whereby they compare all images in a global average space through performing image registration to a template. Voxel-based morphometry (VBM) is one such common method \cite{ashburner2000voxel}, which has been used in countless studies of development, ageing and dementia \cite{matsuda2013voxel, diaz2014brief, busatto2008voxel, padilla2015brain}. More recent methods use Gaussian processes \cite{o2020modelling} to detect diseased brain tissue as outliers against a normative model, fit at each voxel. While these methods have significantly improved understanding of population average patterns of disease \cite{matsuda2013voxel}, they rely on spatial normalisation and therefore lose power at the cortex due to the impact of cortical heterogeneity \cite{dalca2019learning,glasser2016multi}. This also means that they are not tuned to detect features of disease specific to the individual, which are nonetheless important for diagnosis or prognosis.

To address these limitations, recent studies have started to apply deep learning methods to brain imaging datasets. Deep learning is state-of-the-art for many image processing tasks \cite{goodfellow2016deep}, and has shown strong promise for brain imaging applications such as healthy tissue and lesion segmentation \cite{chen2018voxresnet,de2015deep,kamnitsas2017ensembles,rajchl2018neuronet}. Importantly, by design it can work independently of any requirement for spatial normalisation. However, deep learning methods do not, by default, return explanations of the reasoning behind their predictions, leading to them traditionally being referred to as "black box" models. 

More recently, several approaches have been developed to make these networks more interpretable through identifying class-relevant features for a particular input. These include post-hoc saliency based methods, designed to detect which features of a specific image contribute most strongly to a class prediction. These typically analyse the gradients or activations of the network, with respect to a given input image, and include approaches such as Gradient-weighted Class Activation Mapping (Grad-CAM) \cite{selvaraju2017grad}, SHAP \cite{lundberg2017unified}, DeepTaylor \cite{montavon2016deeptaylor}, integrated gradients \cite{sundararajan2017axiomatic}, guided backpropagation (backprop) \cite{springenberg2014striving}, and Layer-wise backpropagation (LRP) \cite{bach2015pixel}. In addition, perturbation methods such as occlusion \cite{zeiler2014visualizing} change or remove parts of the input image to generate heatmaps, by evaluating its effect on the classification prediction. 

Such methods have now been applied in various medical imaging applications including in MRI and Positron Emission Tomography (PET) imaging datasets for Alzheimer's (AD) \cite{li2019deep, ding2019deep, eitel2019testing, bohle2019layer} and Multiple Sclerosis (MS) \cite{eitel2019uncovering} classification. However, while in principle, these methods can be applied to detect features from individual images, the results are typically low resolution and noisy, which makes them hard to interpret. Often this leads to studies estimating a group average to aggregate results across individuals, and boost signal to noise to make stable population-wide inferences \cite{bohle2019layer, eitel2019testing}. This loses individual specificity, and since these feature attribution (FA) methods often detect similar features in both healthy and disease groups, it is difficult to interpret the results.

In addition, since these FA methods are applied to a CNN following training, their power is limited by the constraints of the network they are applied to. In particular, applying a method post-hoc to a classification network has repeatedly been shown to be insufficient, as such networks need only focus on the most consistent or discriminative features, sufficient to accurately predict each class. This is particularly important in medical imaging where diagnosis and treatment rely on comprehensive capture of all features of disease \cite{li2019deep, ding2019deep, eitel2019testing, bohle2019layer, eitel2019uncovering, baumgartner2018visual}. For example, when applying LRP and guided backprop to brain MRI, it was found that while they were able to detect homogeneous brain structures such as the hippocampus, they were unable to detect heterogeneous structures such as cortical folds \cite{bohle2019layer, eitel2019testing}. 

For these reasons, new approaches have recently been proposed which use generative models to translate images from one class to another \cite{isola2017image, zhu2017unpaired}. These provide more comprehensive interpretation, since generative models must capture all relevant features of a population, in order to support synthesis of new images. In Baumgartner et al.  \cite{baumgartner2018visual} for example, a generative model (visual attribution or VA-GAN) was adapted to translate images classed as Alzheimer's (AD) to instead resemble Mild Cognitive Impairment (MCI). However, while this was able to detect more features of disease relative to post-hoc methods, it was still unable to identify much of the phenotypically variable changes (for example heterogeneous patterns of cortical atrophy); see related works for further detail. 

To address these problems in \cite{bass2020icam} we developed ICAM  (Interpretable Classification via disentangled representations and feature Attribution Mapping); this improved on the state-of-the-art FA methods (Table \ref{table:adni}) \cite{selvaraju2017grad, springenberg2014striving, sundararajan2017axiomatic, zeiler2014visualizing, baumgartner2018visual} by building on approaches for image-to-image translation \cite{lee2019drit} to perform feature attribution by disentangling class-relevant \textit{attributes} (attr) from class-irrelevant \textit{content}. Sharp reconstructions are then learnt through use of a Variational Autoencoder (VAE) with a discriminator loss on the decoder (Generative Adversarial Network, GAN). This not only allows classification and generation of an attribution map from the latent space, but also a more interpretable latent space that can visualise differences between and within classes. By sampling the latent space at test time to generate an FA map, we demonstrated its ability to detect meaningful brain variation (Fig.~\ref{fig:adni_comparison}) in 3D brain MRI. 

In this paper, we extend ICAM \cite{bass2020icam} by adding a regression module to enable the network to do regression as well as classification; while past FA methods have been predominantly implemented for classification networks, regression tasks are common in medical imaging, as most diseases lie on a continuous spectrum rather than a binary scale. 
The specific contributions in comparison to previous work \cite{bass2020icam} are as follows:
\begin{itemize}
\item[1] We describe the first framework to implement a translation VAE-GAN network for simultaneous regression and feature attribution.
\item[2] This supports the investigation of heterogeneous, continuous phenotypes such as brain ageing and dementia, specifically allowing for study of features attributed to outlier predictions, to give further insights into the model's reasoning behind these predictions. To our knowledge, this is the first method to provide meaningful explanations in 3D medical imaging regression tasks (with deep learning). 
\item[3] We perform additional experiments, specifically, using the UK Biobank healthy ageing dataset to investigate the latent space for regression tasks. For example, we demonstrate that ICAM can provide explanations for subjects predicted as outliers, and even generate meaningful FA maps when interpolating the attribute latent space between 2 subjects of the same age group. Further, we provide evidence that our translation network consistently changes the class of our input images. Also, we use the dHCP dataset to show that ICAM can detect punctate white matter lesions in preterm babies, without explicit training. 
\end{itemize}

\section{Related works}

\begin{figure}[!b]
  \centering
\makebox[\linewidth]{
	\includegraphics[width=1.0\columnwidth]{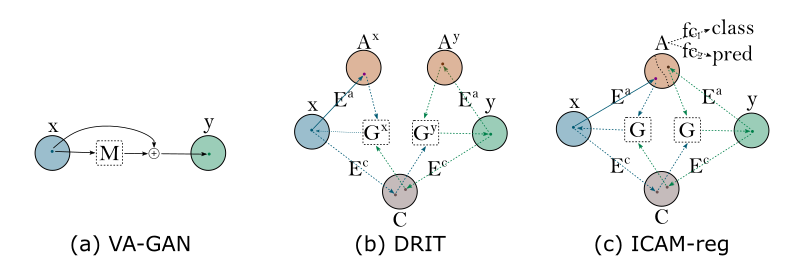}}
\caption{Comparison of domain mapping methods. (a) VA-GAN translates images of domain x to y. (b) DRIT can translate between domains x and y through a shared content space $C$, and separate attribute spaces $A^x$ and $A^y$. (c) $ICAM_{reg}$ uses shared content $C$ and attribute $A$ spaces to translate between domains, which allows classification $f_{C_1}$ and regression $f_{C_2}$ layers to be applied to the attribute space $A$.}
\label{fig:I2I_translation}
\end{figure}

Over recent years, several deep generative approaches to image-to-image translation have emerged \cite{isola2017image,zhu2017unpaired, huang2018multimodal,liu2017unsupervised, jha2018disentangling, lee2019drit}, where these have been applied to many different domains, including medical imaging \cite{baumgartner2018visual, bass2019image, baur2018deep, costa2017end}. Of these, Lee et al. \cite{lee2019drit}, in  particular, developed a domain translation network called DRIT (Fig.~\ref{fig:I2I_translation}b), which constrains translation only to features specific to a class, by encoding separate class-relevant (attribute) and class-irrelevant (content) latent spaces, and employing a discriminator.

Separately, Baumgartner et al. \cite{baumgartner2018visual} developed a conditional GAN-based approach, called VA-GAN, that uses domain translation for feature attribution in medical imaging. In this work,  mappings $M$ were learnt, which translated input 3D MRI brain scans classified with Alzheimer's disease (AD), towards more closely resembling scans with mild cognitive impairment (MCI): an intermediate state between healthy cognition and Alzheimer's disease (Fig.~\ref{fig:I2I_translation}a). This resulted in sharp reconstructions and realistic difference maps that overlap with ground truth outcomes, where available. However, one constraint of VA-GAN is that the approach requires image class labels to be known \textit{a priori} and, in the absence of a latent space, it can only produce a single deterministic output for each image, which limits the modelling of more heterogeneous features.

Accordingly, in our work ICAM \cite{bass2020icam}, we extended upon the intuitions of these models to create one framework which allows simultaneous classification and feature attribution, using a more interpretable model. Compared to VA-GAN and DRIT++ \cite{baumgartner2018visual, lee2019drit}, ICAM uses 2 shared disentangled latent spaces, attribute and content, which encode for class-relevant and class-irrelevant information, respectively. The use of a shared attribute (class) latent space allows the addition of a classification layer (and in this work, also a regression layer) to the network (Fig.~\ref{fig:I2I_translation}c), which enables the network to do classification and visualisation of differences between and within classes. In addition, the rejection sampling module (Fig.~\ref{fig:random_sampling}) checks the class of a randomly sampled attribute latent vector (using the classification layer) to enable feature detection for a single subject during test time, as well as the analysis of the model's mean and variance by sampling the attribute latent space multiple times (in comparison to DRIT++ \cite{lee2019drit}). Other components of ICAM such as a FA map loss, L2 reconstruction loss, and a 3D attribute latent space also improve performance compared to VA-GAN and DRIT++ (as illustrated using ablation studies in \cite{bass2020icam}). Compared to previous works \cite{lee2019drit,baumgartner2018visual, selvaraju2017grad, sundararajan2017axiomatic, springenberg2014striving, zeiler2014visualizing} ICAM demonstrated considerably better feature detection in Alzheimer's and ageing datasets for both consistent (e.g. ventricles and hippocampus) and phenotypically variable (e.g. patterns of cortical atrophy) features of disease (see example in Fig.~\ref{fig:adni_comparison}).

To allow further flexibility, here we extend ICAM with a regression module to enable its application to continuous prediction tasks (Figs.~\ref{fig:biobank_age_pred}, \ref{fig:dhcp_age_prediction}). This also allows further exploration of the latent space (Fig.~\ref{fig:tsne_biobank_reg}, \ref{fig:biobank_interpolation_within}), and outlier subject analysis (Fig.~\ref{fig:biobank_regression}).

\section{Methods}

The goal of the ICAM framework (see Fig.~\ref{fig:network_architecture} for full details of the network architecture) is to perform classification (or regression) with simultaneous feature attribution, by training a VAE-GAN to swap the classes of input images: $x$, $y$; changing only the features of each image which are specific to the target phenotype. The design of the network is outlined in Fig.~\ref{fig:network_translation}, with specific details of the components described below:

\begin{figure}[!hbt]
 \centering
\makebox[\linewidth]{
	\includegraphics[width=1.0\columnwidth]{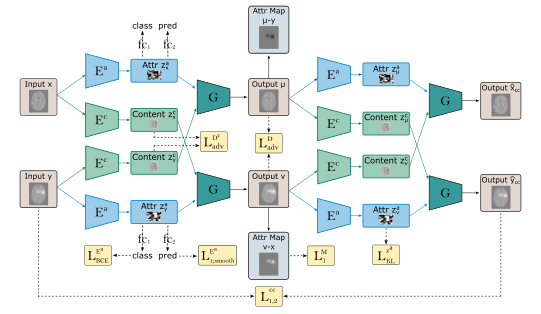}}
\caption{Overview of method. An example of how ICAM performs classification/ regression with attribute map generation for 2 given input images x and y (of class 0 [brain slice without lesions] and 1 [brain slice with simulated lesions], respectively). Note that $L^{D}_{adv}$ is applied to both real and generated images, and that not all losses are plotted (see Equation \ref{eq:objective} for full objective).}
\label{fig:network_translation}
\end{figure}

\subsection{Content and attribute latent spaces} 

To achieve domain disentanglement, two separate latent spaces are encoded: a  \textbf{content encoder} $\{E^c\}$ (latent space $z^c$), whose objective is to encode class-irrelevant (e.g. brain shape) information, and an \textbf{attribute encoder} $\{E^a\}$ (latent space $z^a$), whose objective is to encode all class-relevant features of disease. In both cases, the latent spaces are shared between classes or domains  (i.e. $\{E_c : x \rightarrow C\}, \{E_c : y\rightarrow C\}) $. Note, in what follows, we refer to domain or class interchangeably, in which the same meaning is implied.

For the \textbf{content encoder} $\{E^c\}$, class information is driven out from the latent space $\{C\}$ through training of a discriminator, $\{D^c\}$, with \textbf{class adversarial content loss}:
\begin{equation}
  \begin{aligned}
L_{adv}^{D^c} = \E_{z_x^c}[\log D^c(E^c(x)) + \log(1-D^c(E^c(x)))] \\
+ \E_{z_y^c}[\log D^c(E^c(y)) + \log(1-D^c(E^c(y)))].
  \end{aligned}
\end{equation}

\noindent The goal of the content encoder $\{E_c\}$ is therefore to learn a representation whose domain cannot be distinguished by this discriminator (an approach first proposed by Lee et al., \cite{lee2019drit}). Training is also supported through L2 regularisation, to prevent explosion of gradients, and Gaussian noise (added to the last layer of the encoder) to prevent the latent space vanishing.

For the \textbf{attribute encoder} $\{E^a\}$,  class information is driven \textit{into} the latent space, by appending a fully connected classification layer ($f_{C_1}$) with binary cross entropy loss $L_{BCE}^{E^a}$. In extension from our previous work \cite{bass2020icam}, a \textbf{regression module} $f_{C_2}$) is also added, using another fully connected layer, trained using a smooth L1 loss ($L_{1;smooth}^{E^a}$). 

The training of the attribute latent space is performed using variational inference, through application of a Kullback Leibler (KL) loss $L_{KL}^{z^a}$. This places a Gaussian prior over the latent variables ensuring that the attribute latent space can be sampled, which allows translation of a single subject at test time, and the generation of mean and variance maps via the use of rejection sampling (see below). During training, the prediction modules $f_{C_1}$ and $f_{C_2}$ therefore work to encourage separation of the domains within this latent space $\{A\}$, to support meaningful image translation. Further, to encourage invertible mapping between the image and the latent space,  a \textbf{cyclic reconstruction loss} is added, where a random attribute latent vector $z^a_r$ is sampled from a Gaussian distribution, and reconstructed:
\begin{equation}
L_1^{z^a} = \|E^a(G(E^c(x),z^a_r))-z^a_r\|_1. 
\end{equation}

\begin{figure}[!t]
  \centering
\makebox[\linewidth]{
	\includegraphics[width=0.9\columnwidth]{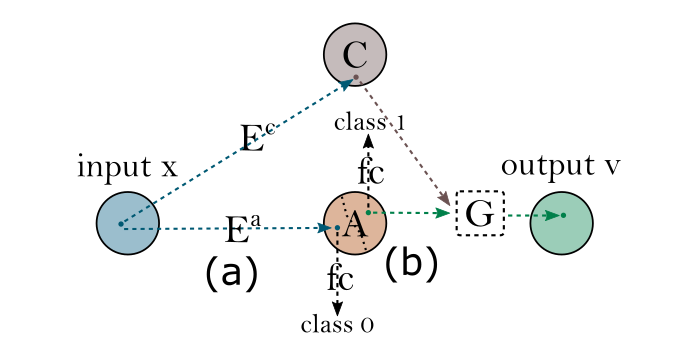}}
\caption{Rejection sampling during training/ testing. Using ICAM, translation can be achieved using a single input image, in addition to translating between 2 images. (a) An input image is encoded into content and attribute spaces, and is passed through the classifier to identify its class (0 in this example). (b) Attribute space A is then randomly sampled until a random vector of the opposite class is sampled (class 1 in this case), by checking its class using the classifier. The newly sampled vector is passed to the generator along with the encoded content space to achieve translation between class 0 and 1. At test time, it is possible to sample the attribute latent space multiple times to get mean and variance FA maps.}
\label{fig:random_sampling}
\end{figure}

Finally, disentanglement is further encouraged through \textbf{rejection  sampling} of the attribute latent space during training by checking the class of a randomly sampled vector using the attribute encoder's classification layer (Fig.~\ref{fig:random_sampling}). Samples are rejected if they belong to the wrong class, which stabilises optimisation of translation by passing the generator samples of the expected class, and allows the generation of mean and variance FA maps at test time (see an example in Fig.~\ref{fig:adni_comparison}). This visualisation approach has previously not been possible in other feature attribution methods, as they do not have a latent space with a classification or regression layer.

\newpage
\subsection{Generation and feature attribution}
Image translation and generation of disease (or FA) maps is supported through the training a \textbf{generator} $\{G\}$, which learns to synthesise images conditioned on both the content and attribute latent spaces $(G: \{z_x^c, z_x^a\} \rightarrow \hat{x}), (G: \{z_y^c, z_y^a\} \rightarrow \hat{y})$, as well as to translate between these domains. It achieves this by swapping the content latent space: $(G: \{z_y^c, z_x^a\} \rightarrow \mu), (G: \{z_x^c, z_y^a\} \rightarrow v)$, which is made possible since this space is class invariant. Training of the generator is supported by optimisation of a \textbf{domain discriminator} $\{D\}$ with two losses: a) a domain adversarial loss, $L^{D}_{adv}$ which seeks realistic image generation by minimising the differences between translated (fake) and real images; and b) a binary cross entropy classification loss, $L^{D}_{BCE}$, which seeks optimal classification of the two domains following translation.

\begin{figure*}[!htb]
  \centering
\makebox[\linewidth]{
	\includegraphics[width=0.9\textwidth]{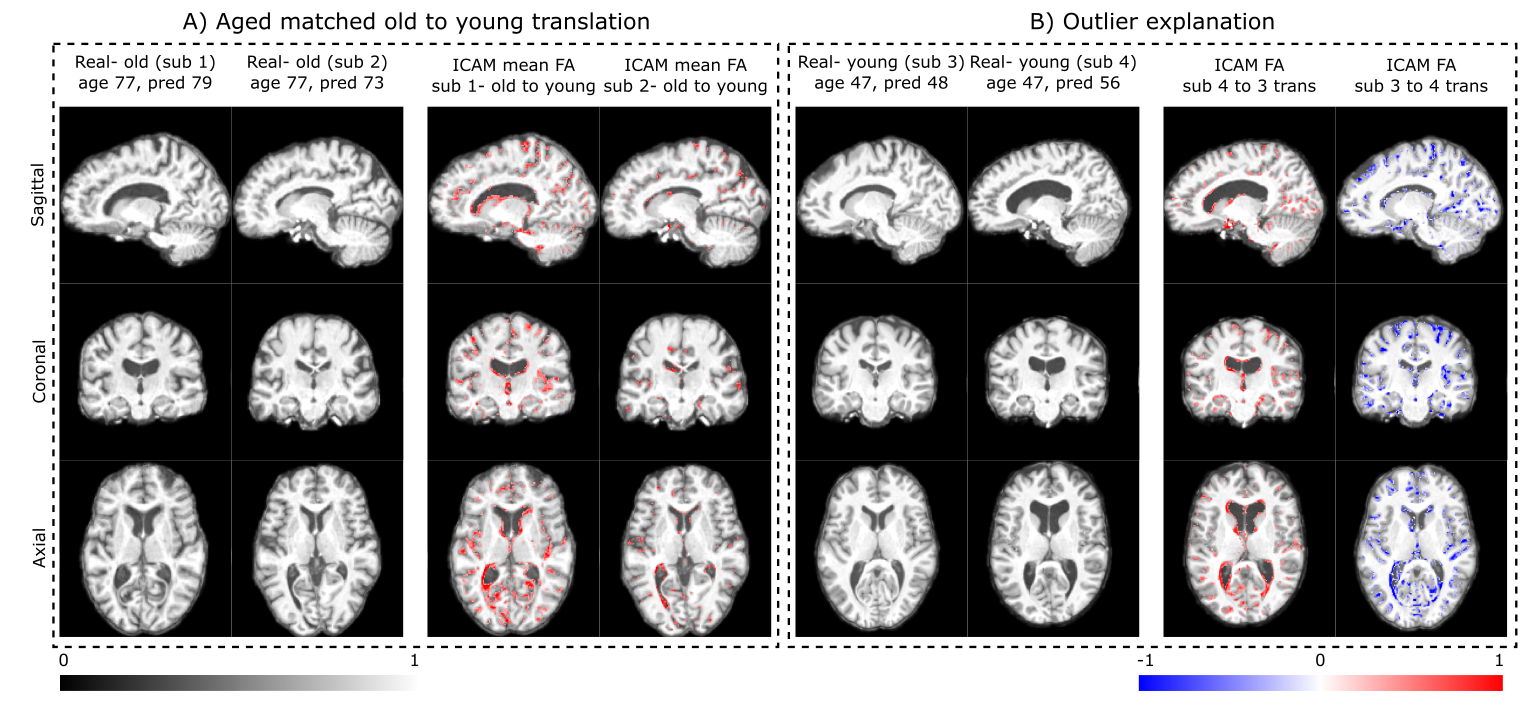}}
\caption{UK Biobank regression results. Here we show examples of the FA map results of our regression model $ICAM_{reg}$ on 4 subjects, with actual age, alongside a predicted age by $ICAM_{reg}$, where FA corresponds to translation of the scans from  A) old to young (using rejection sampling to generate mean FA maps) and B) young to young (by translating between subject 3 and 4 to generate single FA maps). In each case FA maps return explanations for outlier predictions for age-matched subjects.}

\label{fig:biobank_regression}
\end{figure*}

To visualise differences between the translated images $\{v, \mu\}$ and the original images $\{x, y\}$, we use a \textbf{feature attribution map} $\{M\}$. This aims to retain only class-related differences between two images (or two locations in the attribute latent space) by subtracting the content from the translated output $(\{M_{x} = v-x\}, \{M_{y} = \mu-y\})$. Generation is regularised through an L1 loss ($L_1^{M} = \| M(\;)\|_1,$) which encourages $\{M\}$ to reflect a small feasible map, which leads to a realistic translated image. 

Finally, to further facilitate image generation, we apply an L1 and L2 loss to the reconstructed images $\{\hat{x}, \hat{y}\}$ ($L_{1,2}^{rec}$), and the cyclically reconstructed images $\{\hat{x_{cc}}, \hat{y_{cc}}\}$ ($L_{1,2}^{cc}$). The cycle consistency term also allows training with unpaired images.

\begin{equation}
  \begin{aligned}
L_{1,2}^{rec} = \E_{x,y}[\|G(E^c(x), E^a(x))-x\|_{1,2} \\ + \|G(E^c(y), E^a(y))-y\|_{1,2}],
  \end{aligned}
\end{equation}

\begin{equation}
  \begin{aligned}
L_{1,2}^{cc} = \E_{x,y}[\|G(E^c(v), E^a(\mu))-x\|_{1,2} \\ + \|G(E^c(\mu), E^a(v))-y\|_{1,2}].
  \end{aligned}
\end{equation}

This means the \textbf{full objective function}\footnote{See Appendix for full details on training and $\lambda$ values. All of the important components of the network were evaluated through an ablation study in Bass et al., \cite{bass2020icam}.} of our network is:
\begin{equation}
  \begin{aligned}
\min_{G, E^c, E^a} \max_{D, D^c}  \lambda_{D^c} L_{adv}^{D^c} + \lambda_{D} L^{D}_{adv}  + \lambda_{D_{BCE}} L^{D}_{BCE} \\ + \lambda_{BCE} (L_{BCE}^{E^a} + L_{1;smooth}^{E^a}) + \lambda_{KL} L_{KL}^{z^a}+ \lambda_{M} L_1^{M}
+\lambda_{z^a} L_1^{z^a}\\  + \lambda_{rec} (L_1^{rec} + L_1^{cc} + L_2^{rec} +  L_2^{cc}) .
  \end{aligned}
  \label{eq:objective}
\end{equation}

\section{Results}

We evaluate the performance of ICAM through studies on three datasets to test our regression model on 1) an age prediction task (using healthy ageing data from UK Biobank), 2) prediction of birth age (or degree of prematurity) for the dHCP cohort of neonates, born between 23-44 weeks gestation and scanned at term equivalent age; and 3) prediction of the MMSE cognitive test score from the ADNI cohort.

\begin{figure*}[!hbt]
  \centering
\makebox[\linewidth]{
	\includegraphics[width=1.0\textwidth]{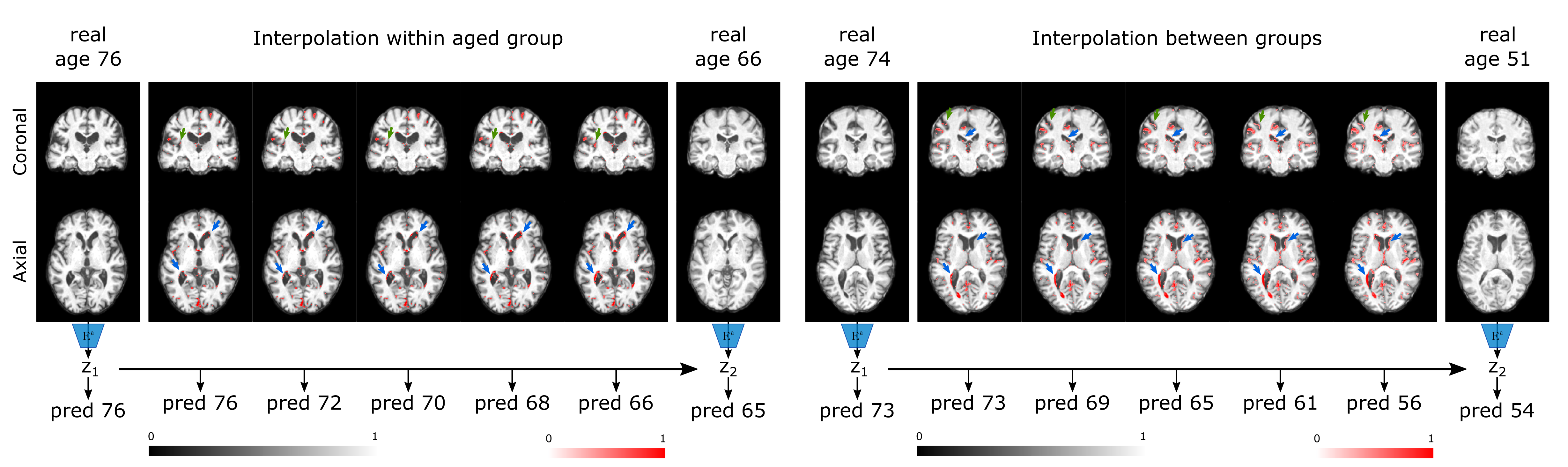}}
\caption{Biobank interpolation between and within groups. Here, we show an example of interpolation of the attribute latent space, with the corresponding FA maps for each vector. We overlay the interpolated FA maps on the original image, with red maps indicating an increase pixel intensity. We first encode each image to its attribute latent space (using $E^a$), and get an age prediction. We then linearly interpolate between these two spaces, and get an age prediction and FA map for each vector. We demonstrate that our $ICAM_{reg}$ model can successfully achieve interpolation between and within groups (i.e. within the aged group, and between the aged and young groups). We find that we get both smoothly interpolated FA maps, and interpolated age predictions between two subjects. The green arrows point to the cortex, and blue arrows point to the ventricles.}
\label{fig:biobank_interpolation_within}
\end{figure*}

\subsection{Brain Age Prediction for the UK Biobank cohort} 

\subsubsection{UK Biobank Dataset}

In this first experiment, we used T1 MRI data from the UK Biobank \cite{alfaro2018image, miller2016multimodal}, a collection of brain imaging data of mostly healthy subjects between the ages of 45-80 years, to map heterogeneous patterns of brain ageing within individuals. 
T1 image processing (see also \cite{alfaro2018image}) involved bias correction using FAST \cite{zhang2001segmentation}, brain extraction using BET \cite{smith2002fast} and linear registration to MNI space, using FLIRT \cite{jenkinson2002improved}. The input into the networks was resized to $128\times160\times128$ voxels, and normalised in range $[0,1]$, per subject.
For our classification experiments we used 11,735 MRI volumes, with the young subjects (45-60 years, average age $54.6 \pm 3.4$ years) separated into training, validation, and testing sets with 6706, 373 and 372 in each. The older subjects (70-80 years, average age $73.0 \pm 2.2$ years) were separated into training, validation, and testing sets with 3856, 214 and 214 in each, respectively. 
For our regression experiments, we used a larger dataset (i.e. all available subjects) of 21,388 subjects, between 45-80 years old. We also split the subjects into older and younger groups, as ICAM requires a minimum of 2 classes for training. Here, subjects corresponding to the young class (45-65 years, average age $57.6 \pm 4.8$ years) were separated into training, validation, and testing sets with sizes 10715, 595 and 595 respectively. The older subjects (65-80 years, average age of $70.0 \pm 3.3$ years) were separated into training, validation, and testing sets with sizes 8535, 474 and 474, respectively. 

\subsubsection{UK Biobank Results}

In our previous work \cite{bass2020icam}, we compared ICAM to VA-GAN for AD to MCI feature attribution, and found that ICAM was able detect more variability in the ventricles, cortex, and in the hippocampus. Further, we found that ICAM is better able to change the shape of relevant brain structures, while VA-GAN is only able to alter the pixel intensities.

In this work, to demonstrate more conclusively whether translation by ICAM and VA-GAN fully changes the image class, we trained an independent classifier (with the same architecture as the ICAM attribute encoder); this was trained on the UK Biobank age classification task, using the 3D T1 MRI images, and was tested on outputs generated by ICAM and VA-GAN (keeping to the same train and test examples as before). Results (Table~\ref{table:biobank_generation}) show that classification with images generated by ICAM performs slightly worse than the real data (82.2\% compared to 93.8\%), which is to be expected in a complex 3D generation task. By contrast, VA-GAN outputs perform much worse (12.2\%). Note that because VA-GAN can only translate in one direction, it has only 1 result in the table. 

\begin{table}[ht]
\small
  \caption{Biobank generation experiment comparing accuracy score for classification (young vs old) of real, ICAM generated, and VA-GAN generated data. Note that because VA-GAN can only do old to young translation, it has only 1 result in the table. }
  \label{table:biobank_generation}
  \centering
  \begin{tabular}{lll}
    \toprule
    \cmidrule(r){1-2}
    Dataset & Accuracy - young & Accuracy - old \\
    \midrule
    Real &  \bf 0.938 & 0.859 \\
    ICAM (translated) &  0.822 &  \bf 0.865 \\
    VA-GAN (translated) &  0.122 & N/A \\

    \bottomrule
  \end{tabular}
\end{table}

\begin{figure}[!htb]
  \centering
\makebox[\linewidth]{
	\includegraphics[width=0.7\columnwidth]{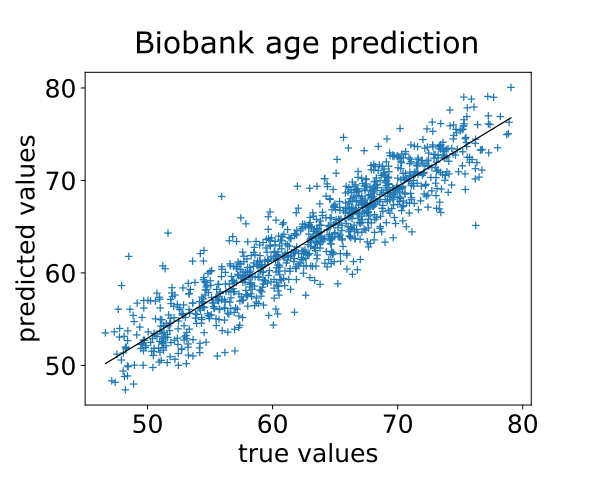}}
\caption{Biobank age prediction on the test dataset using $ICAM_{reg}$. The age prediction error is $2.20 \pm 1.86$ MAE.}
\label{fig:biobank_age_pred}
\end{figure}

\begin{figure*}[!htb]
  \centering
\makebox[\linewidth]{
	\includegraphics[width=0.8\textwidth]{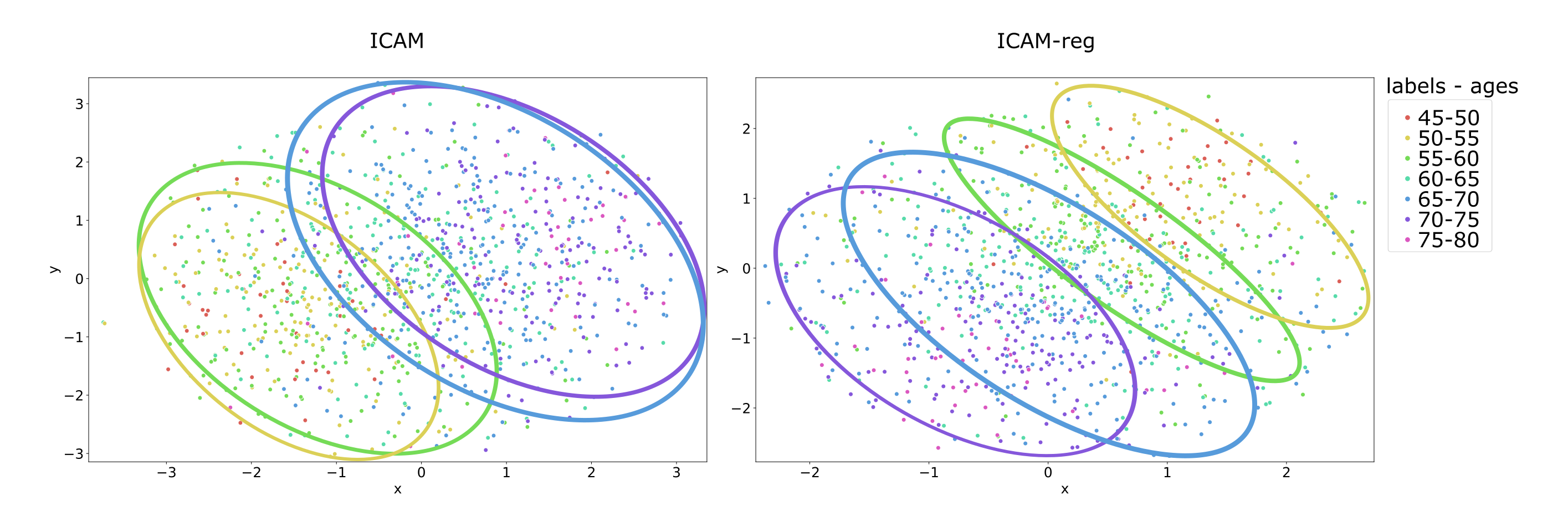}}
\caption{Biobank tSNE comparison. Here we show a comparison of a tSNE plot for $ICAM$ and $ICAM_{reg}$ for a breakdown of ages. We find that $ICAM_{reg}$ has better separation in the latent space compared to $ICAM$.}
\label{fig:tsne_biobank_reg}
\end{figure*}

Next, we trained ICAM's regression layer to predict ages of the MRI brain scans. We found an age prediction error of $2.20 \pm 1.86$ MAE (Fig.~\ref{fig:biobank_age_pred}). In addition, we found that using our regression model, we can generate FA maps that explain the prediction results, including outliers. For example in Fig.~\ref{fig:biobank_regression} A), FA maps of two subjects, scanned at 77 years, and translated to resemble the younger age class, indicate greater age related changes for brain areas commonly associated with ageing (e.g. ventricular and cortical atrophy) in subject 1 (which is predicted as older - 79) relative to subject 2 (which is predicted as younger - 73). In B), 2 subjects from the young group are directly compared by translating between them. Here, subject 4 is predicted to be much older than their true age (predicted=56; true=47 years); whereas, subject 3 has predicted age 49, close to their true age (47). Evidence for the outlier prediction of subject 4 is presented through the translation, indicating the presence of larger ventricles, hippocampal atrophy and cortical shrinking (relative to the more typical presentation of subject 3). To further illustrate this correlation, we then performed a Pearson's correlation test (on test subjects) between the predicted or true ages and the mean of the corresponding FA maps. We found a stronger positive correlation ($p<0.0001$) using the predicted ages rather than the true ages (0.306 vs 0.255, respectively, for old subjects). This indicates that predicted age is a better predictor of the features related to age, than the true age. This is because very old subjects need to be changed more drastically than younger subjects, and thus have higher mean FA maps.

Finally, we investigated the improvement in separation of the model's latent space afforded through regression (Fig.~\ref{fig:tsne_biobank_reg}, tSNE latent space comparison). This result is further underlined in Fig.~\ref{fig:biobank_interpolation_within}, which shows clearly that interpolation between images of two different ages smoothly translates both predicted ages and feature attribution maps, for the generated images.

\subsection{dHCP experiments} 

\begin{figure}[!htb]
  \centering
\makebox[\linewidth]{
	\includegraphics[width=0.9\columnwidth]{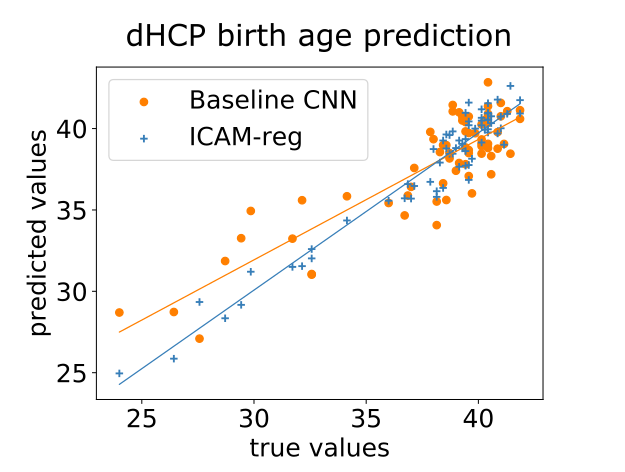}}
\caption{dHCP birth age prediction on the test dataset. The age prediction MAE is $0.806 \pm 0.634$ and $1.525 \pm 1.160$, for ICAM and the baseline network, respectively. The Spearman correlation coefficient is $0.873$ and $0.695$ ($p<0.0001$), for ICAM and the baseline network, respectively. }
\label{fig:dhcp_age_prediction}
\end{figure}

\begin{figure*}[!hbt]
  \centering
\makebox[\linewidth]{
	\includegraphics[width=0.9\textwidth]{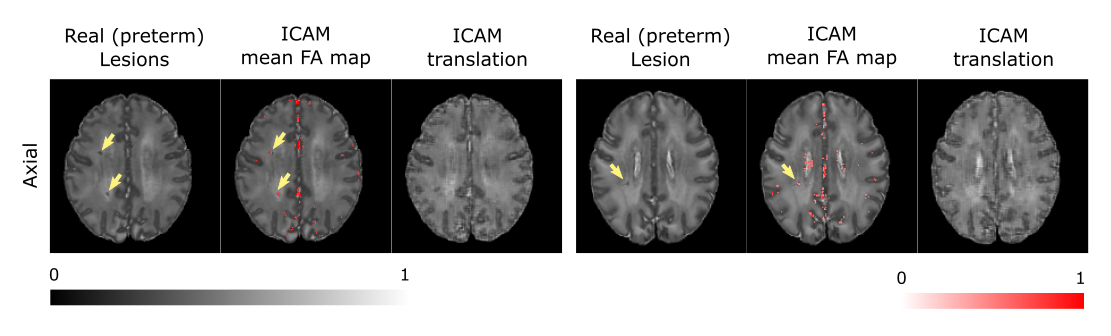}}
\caption{dHCP results. Here we show detection of punctate white matter lesions (yellow arrows) on previously unseen images by $ICAM_{reg}$.}
\label{fig:dhcp_lesions}
\end{figure*}

\subsubsection{dHCP Dataset}

In this experiment we sought to demonstrate that ICAM can work well for prediction of challenging phenotypes, and detection of focal lesions, from relatively small, heterogenous, datasets. We used 699 3D T2 MRI scans from the dHCP \cite{makropoulos2018developing, fitzgibbon2020developing}: an open data set of multimodal brain scans acquired from preterm and term neonates. Here, preterm is defined as birth prior to 37 weeks gestational age (GA), where some preterm neonates were scanned twice: at birth and at term equivalent age. The data set includes 143 preterm images (class 1, mean gestation age at birth: $31.8 \pm 3.85$ weeks, mean post-menstrual age at scan: $41.0 \pm 1.99$ weeks) and 556 term controls (class 0, mean age at birth:  $40.0 \pm 1.27$ week, mean post-menstrual age at scan: $41.4 \pm 1.74$). In this experiment ICAM was trained to classify between preterms and terms, and predict birth age from the term age scan (i.e. scans acquired after 37 weeks post-menstrual). Examples were split into train, validation and test sets according to a 446:55:55 split (for term subjects) and 115:14:14 split (for preterm subjects). 

Image pre-processing involved using diffeomorphic multi-modal (T1w/T2w) registration (ANTs SyN) to estimate nonlinear transforms to a 40 week template from the extended atlas \cite{schuh2018unbiased, avants2011reproducible, fitzgibbon2020developing}. This was necessary to allow the network to train, since without this step, it is suspected that the network was challenged by stark changes in image appearance, which are typically observed for neonatal cohorts, and caused by rapid tissue maturation; this was further confounded by the relatively small and imbalanced nature of the data set. For related reasons (to preserves age-related tissue maturational differences), images were rescaled to [0,1] by normalising across the intensity range of the entire group. Images were then brain extracted (using blurred masks), and CSF, ventricles and the skull were removed in order to focus the attention of the model on brain tissue differences between the groups. 

\subsubsection{dHCP Results}

To compare against a baseline CNN network, we trained a CNN network (with same architecture as $E^a$) for regression using a smooth L1 loss, trained similarly to ICAM (see Appendix for details on training).
Results are shown in Figs. \ref{fig:dhcp_age_prediction} and \ref{fig:dhcp_lesions}. 

We report a birth age prediction MAE of $0.806 \pm 0.634$ vs $1.525 \pm 1.160$ for $ICAM_{reg}$ and the baseline CNN, respectively (Fig.~\ref{fig:dhcp_age_prediction}). In addition, we report a higher correlation coefficient for ICAM (Spearman correlation test, $p<0.0001$, $0.873$ and $0.695$ correlation coefficient for ICAM and the baseline network, respectively.). 

In our qualitative analysis we tested the $ICAM_{reg}$ model on previously unseen images of subjects with punctate white matter lesions (PWML), which are commonly seen in preterm babies \cite{tusor2017punctate}, to test whether these would be detected by ICAM. In this experiment, ICAM is trained to predict birth age, rather than explicitly detect PWML. The results in Fig.\ref{fig:dhcp_lesions} (yellow arrows pointing at the lesions) demonstrate that ICAM successfully and consistently detects these lesions.

\subsection{ADNI experiments: Ground-truth evaluation of feature attribution maps} 

In the final experiment, we demonstrate the performance of ICAM's feature attribution against ground truth maps of disease progression estimated for AD to MCI conversion using the ADNI dataset and extend \cite{bass2020icam} to explore modelling regression of MMSE scores.

\subsubsection{ADNI Dataset}

The data used in this study was obtained from the Alzheimer’s Disease Neuroimaging Initiative (ADNI) database (adni.loni.usc.edu), first launched in 2003, and led by Principal Investigator Michael W. Weiner, MD \cite{jack2008alzheimer}. We used 1,053 3T T1 MRI images of AD and MCI patients with average age of $74.95 \pm 8.1$ (for AD subject) and  $72.26 \pm 7.9$ (for MCI subjects). For AD/MCI test subjects the average age was $73.47 \pm 7.2$. We split the dataset into AD and MCI classes, with 257 and 674 volumes used in training, respectively. For testing, 61 subjects who converted from MCI to AD (i.e. paired data) were used. A further 61 conversion subjects were used for validation. Each of these subjects were scanned before and after conversion to AD. Using these two scans ground truth disease maps were generated through rigid alignment and subtraction. All disease and FA maps were masked to ensure that the returned normalised cross correlation (NCC) values reference brain tissue only.

MRI volumes were pre-processed with N4 bias correction \cite{tustison2010n4itk} implemented in SimpleITK, brain extracted using Freesurfer \cite{segonne2004hybrid} and rigidly registered to the MNI space using Niftyreg \cite{modat2012inverse}. Images were normalised in range $[-1, 1]$ per subject, and resized to $128\times160\times128$ voxels. 

\subsubsection{ADNI Results}

We compared our proposed approach against a range of baselines in our experiments. For a fair comparison, we use the same training, validation and testing datasets. In particular, we compared against Grad-CAM, guided Grad-CAM \cite{selvaraju2017grad}, guided backprop \cite{springenberg2014striving}, Integrated gradients \cite{sundararajan2017axiomatic}, Occlusion \cite{zeiler2014visualizing}, and VA-GAN \cite{baumgartner2018visual}. These methods were applied to a simple 3D ResNet. We compared against two variations of our network, $ICAM$, and $ICAM_{reg}$. See Appendix for full details on comparison methods.

Following the approach of Baumgartner et al. \cite{baumgartner2018visual}, experiments were performed to predict AD vs MCI classification. In extension to our previous work \cite{bass2020icam}, ICAM's regression module was used to predict the MMSE score, a test used to help diagnose Alzheimer's disease (Scores of 25-30 is considered normal cognitive, 21-24 is mild dementia, 10-20 is moderate dementia, and 9 or lower is severe dementia). Here, MCI and AD training sets had an average MMSE score of $27.75 \pm 2.61$ and $23.00 \pm 2.63$, respectively. The MCI and AD test and validation sets had a similar average MMSE score of $26.80 \pm 2.95$ and $ 23.93 \pm  4.33$, respectively.

Results in (Table~\ref{table:adni}, Fig.~\ref{fig:adni_comparison}) show that all versions of ICAM outperform VA-GAN, and other FA baseline methods (occlusion, integrated gradients, Grad-CAM, guided Grad-CAM and guided backprop) with respect to predicting patterns of brain atrophy, which compare well to the ground truth. Specifically, the NCC metric is increased (Table~\ref{table:adni}) and the returned FA maps detect much greater proportions of atrophy (for cortex and hippocampus) and ventricular expansion, relative to the ground truth (Fig.~\ref{fig:adni_comparison}). At the same time, regression of the MMSE score returns MMSE prediction of $2.82 \pm 2.14$ mean absolute error, which is comparable to the standard deviation of the MMSE scores of the test set ($3.64$). Importantly, we cannot compare to VA-GAN or other FA methods, as they cannot be normally applied to regression tasks. 

\begin{table}[ht]
\small
  \caption{ADNI experiments comparing baselines with ICAM. Networks are compared using normalised cross correlation (NCC) between the absolute values of the attribution maps and the ground truth masks. The positive NCC (+) compares the lesion mask to the attribution map when translating between class 0 (MCI) to 1 (AD), and vice versa for the negative NCC (-). Values reported are the mean and standard deviation across the test subjects.}
  \label{table:adni}
  \centering
  \begin{tabular}{llll}
    \toprule
    \cmidrule(r){1-2}
    Network & NCC (-) & NCC (+)  \\
    \midrule
    Guided Grad-CAM \cite{selvaraju2017grad} & $0.244 \pm 0.047$ & $0.339 \pm 0.068 $    \\
    Grad-CAM \cite{selvaraju2017grad} & $0.321 \pm 0.059$ & $0.461 \pm 0.086 $    \\
    Occlusion \cite{zeiler2014visualizing} & $0.360 \pm 0.037$ &   $0.354 \pm 0.057 $  \\
    Integrated gradients \cite{sundararajan2017axiomatic} & $0.378 \pm 0.064 $ &  $0.404 \pm 0.059 $   \\
    Guided backprop \cite{springenberg2014striving} & $0.541 \pm 0.054$ & $0.532 \pm 0.052 $    \\
    VA-GAN \cite{baumgartner2018visual} & $0.653\pm 0.142$  &  N/A      \\
    $ICAM_{reg}$ &  0.655 $\pm$ 0.086 &  0.611 $\pm$ 0.059    \\
    $ICAM$ & \bf 0.683 $\pm$ 0.097 & \bf 0.652 $\pm$ 0.083    \\
    \bottomrule
  \end{tabular}
\end{table}

\section{Discussion}

In our previous work \cite{bass2020icam} we developed a novel framework, ICAM, for classification with feature attribution, and showed that it outperforms state-of-the-art feature attribution methods on classification tasks for individual subject feature detection. In this work, we extended ICAM to include a regression module, $ICAM_{reg}$, that can work on regression tasks. We performed extensive experiments that show $ICAM_{reg}$ is able of successfully performing regression and demonstrated that meaningful explanations can be generated alongside predictions. To our knowledge, this is the first generative method that can provide regression predictions alongside a subject-specific explanation for the prediction. 

Specifically, in our experiments we sought to test whether, when trained on a large dataset (UK Biobank), $ICAM_{reg}$ can learn to disentangle its attribute latent space, so as to support meaningful interpolation between images, and generate subject-specific explanations for outlier predictions. We then set out to show that $ICAM_{reg}$ can also work on much smaller and more heterogeneous datasets (dHCP, ADNI), while continuing to detect relevant features not explicitly defined during training, i.e. white matter lesions in the dHCP neonatal data; and predict clinically relevant phenotypes such as age at birth (dHCP) and cognitive test scores (ADNI).

Through experiments on UK Biobank, we demonstrate that ICAM is much better able to change the class of input images than VA-GAN \cite{baumgartner2018visual} (Table \ref{table:biobank_generation}), which shows that while VA-GAN is only able to slightly modify the images by changing pixel intensities in order to generate FA maps, ICAM can drastically change the input image in order to change its class, and thus also generate more reliable FA maps. In separate experiments, we show that brain age prediction by $ICAM_{reg}$ ($2.20\pm1.86$ MAE, Fig.~\ref{fig:biobank_age_pred}) performs highly competitively relative to other deep learning methods trained on age prediction in UK Biobank (with reported test MAE scores of $2.14 \pm 0.05$ \cite{peng2021accurate}, and $4.006$ \cite{jonsson2019brain}). Alongside the age prediction, we find that ICAM can provide meaningful and individual explanations for old and young classification, as well as outlier predictions (Figs.~\ref{fig:biobank_regression}, \ref{fig:biobank_interpolation_within}). The meaningfulness of this prediction was emphasised through estimating Pearson's correlation, between both the real age and FA maps, and the predicted age and the FA maps. We found that predicted age was more strongly correlated with detected features, suggesting predicted age is a better indicator for FA map generation. We also demonstrated that our regression model has a more interpretable latent space than our previous model \cite{bass2020icam}, through use of a tSNE comparison (Fig.~\ref{fig:tsne_biobank_reg}), and demonstrated interpolation of the latent space between groups and within groups (Fig.~\ref{fig:biobank_interpolation_within}).

In our dHCP experiments, we compared our regression model to a baseline CNN that has the same architecture as our attribute encoder and found that $ICAM_{reg}$ performs better than the baseline CNN on birth age prediction (Fig.~\ref{fig:dhcp_age_prediction}). At the same time, the model returns subject specific FA explanations of the predictions, which consistently detect punctate white matter lesions, within individuals (a known feature of preterm birth, Fig.~\ref{fig:dhcp_lesions}). These are detected despite stark changes in image intensity and appearance over this neonatal period. 

For ADNI we show that $ICAM_{reg}$ can predict cognitive scores related to Alzheimer's (MMSE scores), and still provide meaningful FA map explanations that are better than baseline methods, though slightly worse than our classification model (disease map comparisons, Fig.~\ref{fig:adni_comparison}, Table \ref{table:adni}). We note that this is a very challenging prediction task, as the MMSE scores are not a complete indicator for Alzheimer's, but rather an additional factor that clinicians can use to help with the diagnosis. This could lead to the loss function being less optimal, and thus resulting in reduced NCC scores. It is possible that NCC results could be improved through additional hyperparameter optimisation, or using combined metrics of cognitive tests that are also used for AD and MCI diagnosis, e.g. Addenbrooke’s Cognitive Examination–Revised (ACE-R) and Montreal Cognitive Assessment (MoCA) \cite{tsoi2015cognitive}.

Finally, there are still several challenges in using ICAM, including applying it to small datasets with imbalanced classes, e.g. in the dHCP experiments we observed that the FA maps were noisy relative to results using the UK Biobank. This is not unexpected given the vast difference in sizes of the dataset; however, it confounds  interpretation of the FA maps. In addition, we found that while ICAM worked well for rigidly aligned ADNI and Biobank datasets, it did not work well in early experiments for the dHCP dataset when they were rigidly aligned. We instead used non-linearly aligned data in the experiments shown in this paper. This could be due to vast differences in tissue intensities and shape for neonates, as they change very rapidly at an early age. Furthermore, we found that the latent space is not completely separated even in larger datasets (Fig.~\ref{fig:tsne_biobank_reg}). These challenges could be addressed in future work via application of GAN augmentation techniques \cite{karras2020training} to increase training data for smaller datasets, and latent space clustering strategies to further encourage disentanglement of rare classes \cite{sohoni2020no}. This would also help in training imbalanced datasets.

\bibliographystyle{IEEEtran}
\bibliography{mybibliography}

\begin{thebibliography}{10}
\providecommand{\url}[1]{#1}
\csname url@samestyle\endcsname
\providecommand{\newblock}{\relax}
\providecommand{\bibinfo}[2]{#2}
\providecommand{\BIBentrySTDinterwordspacing}{\spaceskip=0pt\relax}
\providecommand{\BIBentryALTinterwordstretchfactor}{4}
\providecommand{\BIBentryALTinterwordspacing}{\spaceskip=\fontdimen2\font plus
\BIBentryALTinterwordstretchfactor\fontdimen3\font minus
  \fontdimen4\font\relax}
\providecommand{\BIBforeignlanguage}[2]{{%
\expandafter\ifx\csname l@#1\endcsname\relax
\typeout{** WARNING: IEEEtran.bst: No hyphenation pattern has been}%
\typeout{** loaded for the language `#1'. Using the pattern for}%
\typeout{** the default language instead.}%
\else
\language=\csname l@#1\endcsname
\fi
#2}}
\providecommand{\BIBdecl}{\relax}
\BIBdecl

\bibitem{cullen2019polygenic}
H.~Cullen, M.~L. Krishnan, S.~Selzam, G.~Ball, A.~Visconti, A.~Saxena, S.~J.
  Counsell, J.~Hajnal, G.~Breen, R.~Plomin \emph{et~al.}, ``Polygenic risk for
  neuropsychiatric disease and vulnerability to abnormal deep grey matter
  development,'' \emph{Scientific reports}, vol.~9, no.~1, pp. 1--8, 2019.

\bibitem{iqbal2005subgroups}
K.~Iqbal, M.~Flory, S.~Khatoon, H.~Soininen, T.~Pirttila, M.~Lehtovirta,
  I.~Alafuzoff, K.~Blennow, N.~Andreasen, E.~Vanmechelen \emph{et~al.},
  ``Subgroups of alzheimer's disease based on cerebrospinal fluid molecular
  markers,'' \emph{Annals of Neurology: Official Journal of the American
  Neurological Association and the Child Neurology Society}, vol.~58, no.~5,
  pp. 748--757, 2005.

\bibitem{ross2006neurobiology}
C.~A. Ross, R.~L. Margolis, S.~A. Reading, M.~Pletnikov, and J.~T. Coyle,
  ``Neurobiology of schizophrenia,'' \emph{Neuron}, vol.~52, no.~1, pp.
  139--153, 2006.

\bibitem{glasser2016multi}
M.~F. Glasser, T.~S. Coalson, E.~C. Robinson, C.~D. Hacker, J.~Harwell,
  E.~Yacoub, K.~Ugurbil, J.~Andersson, C.~F. Beckmann, M.~Jenkinson
  \emph{et~al.}, ``A multi-modal parcellation of human cerebral cortex,''
  \emph{Nature}, vol. 536, no. 7615, pp. 171--178, 2016.

\bibitem{kong2019spatial}
R.~Kong, J.~Li, C.~Orban, M.~R. Sabuncu, H.~Liu, A.~Schaefer, N.~Sun, X.-N.
  Zuo, A.~J. Holmes, S.~B. Eickhoff \emph{et~al.}, ``Spatial topography of
  individual-specific cortical networks predicts human cognition, personality,
  and emotion,'' \emph{Cerebral cortex}, vol.~29, no.~6, pp. 2533--2551, 2019.

\bibitem{ashburner2000voxel}
J.~Ashburner and K.~J. Friston, ``Voxel-based morphometry—the methods,''
  \emph{Neuroimage}, vol.~11, no.~6, pp. 805--821, 2000.

\bibitem{matsuda2013voxel}
H.~Matsuda, ``Voxel-based morphometry of brain mri in normal aging and
  alzheimer’s disease,'' \emph{Aging and disease}, vol.~4, no.~1, p.~29,
  2013.

\bibitem{diaz2014brief}
L.~Z. Diaz-de Grenu, J.~Acosta-Cabronero, Y.~F.~V. Chong, J.~Pereira, S.~A.
  Sajjadi, G.~B. Williams, and P.~J. Nestor, ``A brief history of voxel-based
  grey matter analysis in alzheimer's disease,'' \emph{Journal of Alzheimer's
  Disease}, vol.~38, no.~3, pp. 647--659, 2014.

\bibitem{busatto2008voxel}
G.~F. Busatto, B.~S. Diniz, and M.~V. Zanetti, ``Voxel-based morphometry in
  alzheimer’s disease,'' \emph{Expert review of neurotherapeutics}, vol.~8,
  no.~11, pp. 1691--1702, 2008.

\bibitem{padilla2015brain}
N.~Padilla, G.~Alexandrou, M.~Blennow, H.~Lagercrantz, and U.~{\AA}d{\'e}n,
  ``Brain growth gains and losses in extremely preterm infants at term,''
  \emph{Cerebral Cortex}, vol.~25, no.~7, pp. 1897--1905, 2015.

\bibitem{o2020modelling}
J.~O'Muircheartaigh, E.~C. Robinson, M.~Pietsch, T.~Wolfers, P.~Aljabar, L.~C.
  Grande, R.~P. Teixeira, J.~Bozek, A.~Schuh, A.~Makropoulos \emph{et~al.},
  ``Modelling brain development to detect white matter injury in term and
  preterm born neonates,'' \emph{Brain}, vol. 143, no.~2, pp. 467--479, 2020.

\bibitem{dalca2019learning}
A.~Dalca, M.~Rakic, J.~Guttag, and M.~Sabuncu, ``Learning conditional
  deformable templates with convolutional networks,'' in \emph{Advances in
  neural information processing systems}, 2019, pp. 804--816.

\bibitem{goodfellow2016deep}
I.~Goodfellow, Y.~Bengio, and A.~Courville, \emph{Deep learning}.\hskip 1em
  plus 0.5em minus 0.4em\relax MIT press, 2016.

\bibitem{chen2018voxresnet}
H.~Chen, Q.~Dou, L.~Yu, J.~Qin, and P.-A. Heng, ``Voxresnet: Deep voxelwise
  residual networks for brain segmentation from 3d mr images,''
  \emph{NeuroImage}, vol. 170, pp. 446--455, 2018.

\bibitem{de2015deep}
A.~de~Brebisson and G.~Montana, ``Deep neural networks for anatomical brain
  segmentation,'' in \emph{Proceedings of the IEEE conference on computer
  vision and pattern recognition workshops}, 2015, pp. 20--28.

\bibitem{kamnitsas2017ensembles}
K.~Kamnitsas, W.~Bai, E.~Ferrante, S.~McDonagh, M.~Sinclair, N.~Pawlowski,
  M.~Rajchl, M.~Lee, B.~Kainz, D.~Rueckert \emph{et~al.}, ``Ensembles of
  multiple models and architectures for robust brain tumour segmentation,'' in
  \emph{International MICCAI Brainlesion Workshop}.\hskip 1em plus 0.5em minus
  0.4em\relax Springer, 2017, pp. 450--462.

\bibitem{rajchl2018neuronet}
M.~Rajchl, N.~Pawlowski, D.~Rueckert, P.~M. Matthews, and B.~Glocker,
  ``Neuronet: fast and robust reproduction of multiple brain image segmentation
  pipelines,'' \emph{arXiv preprint arXiv:1806.04224}, 2018.

\bibitem{selvaraju2017grad}
R.~R. Selvaraju, M.~Cogswell, A.~Das, R.~Vedantam, D.~Parikh, and D.~Batra,
  ``Grad-cam: Visual explanations from deep networks via gradient-based
  localization,'' in \emph{Proceedings of the IEEE international conference on
  computer vision}, 2017, pp. 618--626.

\bibitem{lundberg2017unified}
S.~M. Lundberg and S.-I. Lee, ``A unified approach to interpreting model
  predictions,'' in \emph{Advances in neural information processing systems},
  2017, pp. 4765--4774.

\bibitem{montavon2016deeptaylor}
G.~Montavon, S.~Bach, A.~Binder, W.~Samek, and K.-R. Muller, ``Deep taylor
  decomposition of neural networks,'' in \emph{International Conference on
  Machine Learning workshop on Visualization for Deep Learning}, 2016.

\bibitem{sundararajan2017axiomatic}
M.~Sundararajan, A.~Taly, and Q.~Yan, ``Axiomatic attribution for deep
  networks,'' in \emph{Proceedings of the 34th International Conference on
  Machine Learning-Volume 70}.\hskip 1em plus 0.5em minus 0.4em\relax JMLR.
  org, 2017, pp. 3319--3328.

\bibitem{springenberg2014striving}
J.~T. Springenberg, A.~Dosovitskiy, T.~Brox, and M.~Riedmiller, ``Striving for
  simplicity: The all convolutional net,'' \emph{arXiv preprint
  arXiv:1412.6806}, 2014.

\bibitem{bach2015pixel}
S.~Bach, A.~Binder, G.~Montavon, F.~Klauschen, K.-R. M{\"u}ller, and W.~Samek,
  ``On pixel-wise explanations for non-linear classifier decisions by
  layer-wise relevance propagation,'' \emph{PloS one}, vol.~10, no.~7, 2015.

\bibitem{zeiler2014visualizing}
M.~D. Zeiler and R.~Fergus, ``Visualizing and understanding convolutional
  networks,'' in \emph{European conference on computer vision}.\hskip 1em plus
  0.5em minus 0.4em\relax Springer, 2014, pp. 818--833.

\bibitem{li2019deep}
H.~Li, M.~Habes, D.~A. Wolk, Y.~Fan, A.~D.~N. Initiative \emph{et~al.}, ``A
  deep learning model for early prediction of alzheimer's disease dementia
  based on hippocampal magnetic resonance imaging data,'' \emph{Alzheimer's \&
  dementia}, vol.~15, no.~8, pp. 1059--1070, 2019.

\bibitem{ding2019deep}
Y.~Ding, J.~H. Sohn, M.~G. Kawczynski, H.~Trivedi, R.~Harnish, N.~W. Jenkins,
  D.~Lituiev, T.~P. Copeland, M.~S. Aboian, C.~Mari~Aparici \emph{et~al.}, ``A
  deep learning model to predict a diagnosis of alzheimer disease by using
  18f-fdg pet of the brain,'' \emph{Radiology}, vol. 290, no.~2, pp. 456--464,
  2019.

\bibitem{eitel2019testing}
F.~Eitel, K.~Ritter, A.~D. N.~I. (ADNI \emph{et~al.}, ``Testing the robustness
  of attribution methods for convolutional neural networks in mri-based
  alzheimer’s disease classification,'' in \emph{Interpretability of Machine
  Intelligence in Medical Image Computing and Multimodal Learning for Clinical
  Decision Support}.\hskip 1em plus 0.5em minus 0.4em\relax Springer, 2019, pp.
  3--11.

\bibitem{bohle2019layer}
M.~B{\"o}hle, F.~Eitel, M.~Weygandt, and K.~Ritter, ``Layer-wise relevance
  propagation for explaining deep neural network decisions in mri-based
  alzheimer’s disease classification,'' \emph{Frontiers in aging
  neuroscience}, vol.~11, p. 194, 2019.

\bibitem{eitel2019uncovering}
F.~Eitel, E.~Soehler, J.~Bellmann-Strobl, A.~U. Brandt, K.~Ruprecht, R.~M.
  Giess, J.~Kuchling, S.~Asseyer, M.~Weygandt, J.-D. Haynes \emph{et~al.},
  ``Uncovering convolutional neural network decisions for diagnosing multiple
  sclerosis on conventional mri using layer-wise relevance propagation,''
  \emph{NeuroImage: Clinical}, vol.~24, p. 102003, 2019.

\bibitem{baumgartner2018visual}
C.~F. Baumgartner, L.~M. Koch, K.~Can~Tezcan, J.~Xi~Ang, and E.~Konukoglu,
  ``Visual feature attribution using wasserstein gans,'' in \emph{Proceedings
  of the IEEE Conference on Computer Vision and Pattern Recognition}, 2018, pp.
  8309--8319.

\bibitem{isola2017image}
P.~Isola, J.-Y. Zhu, T.~Zhou, and A.~A. Efros, ``Image-to-image translation
  with conditional adversarial networks,'' in \emph{Proceedings of the IEEE
  conference on computer vision and pattern recognition}, 2017, pp. 1125--1134.

\bibitem{zhu2017unpaired}
J.-Y. Zhu, T.~Park, P.~Isola, and A.~A. Efros, ``Unpaired image-to-image
  translation using cycle-consistent adversarial networks,'' in
  \emph{Proceedings of the IEEE international conference on computer vision},
  2017, pp. 2223--2232.

\bibitem{bass2020icam}
C.~Bass, M.~da~Silva, C.~Sudre, P.-D. Tudosiu, S.~Smith, and E.~Robinson,
  ``Icam: Interpretable classification via disentangled representations and
  feature attribution mapping,'' \emph{Advances in Neural Information
  Processing Systems}, vol.~33, 2020.

\bibitem{lee2019drit}
H.-Y. Lee, H.-Y. Tseng, Q.~Mao, J.-B. Huang, Y.-D. Lu, M.~Singh, and M.-H.
  Yang, ``Drit++: Diverse image-to-image translation via disentangled
  representations,'' \emph{arXiv preprint arXiv:1905.01270}, 2019.

\bibitem{huang2018multimodal}
X.~Huang, M.-Y. Liu, S.~Belongie, and J.~Kautz, ``Multimodal unsupervised
  image-to-image translation,'' in \emph{Proceedings of the European Conference
  on Computer Vision (ECCV)}, 2018, pp. 172--189.

\bibitem{liu2017unsupervised}
M.-Y. Liu, T.~Breuel, and J.~Kautz, ``Unsupervised image-to-image translation
  networks,'' in \emph{Advances in neural information processing systems},
  2017, pp. 700--708.

\bibitem{jha2018disentangling}
A.~H. Jha, S.~Anand, M.~Singh, and V.~Veeravasarapu, ``Disentangling factors of
  variation with cycle-consistent variational auto-encoders,'' in
  \emph{European Conference on Computer Vision}.\hskip 1em plus 0.5em minus
  0.4em\relax Springer, 2018, pp. 829--845.

\bibitem{bass2019image}
C.~Bass, T.~Dai, B.~Billot, K.~Arulkumaran, A.~Creswell, C.~Clopath,
  V.~De~Paola, and A.~A. Bharath, ``Image synthesis with a convolutional
  capsule generative adversarial network,'' \emph{Medical Imaging with Deep
  Learning}, 2019.

\bibitem{baur2018deep}
C.~Baur, B.~Wiestler, S.~Albarqouni, and N.~Navab, ``Deep autoencoding models
  for unsupervised anomaly segmentation in brain mr images,'' in
  \emph{International MICCAI Brainlesion Workshop}.\hskip 1em plus 0.5em minus
  0.4em\relax Springer, 2018, pp. 161--169.

\bibitem{costa2017end}
P.~Costa, A.~Galdran, M.~I. Meyer, M.~Niemeijer, M.~Abr{\`a}moff, A.~M.
  Mendon{\c{c}}a, and A.~Campilho, ``End-to-end adversarial retinal image
  synthesis,'' \emph{IEEE transactions on medical imaging}, vol.~37, no.~3, pp.
  781--791, 2017.

\bibitem{alfaro2018image}
F.~Alfaro-Almagro, M.~Jenkinson, N.~K. Bangerter, J.~L. Andersson,
  L.~Griffanti, G.~Douaud, S.~N. Sotiropoulos, S.~Jbabdi,
  M.~Hernandez-Fernandez, E.~Vallee \emph{et~al.}, ``Image processing and
  quality control for the first 10,000 brain imaging datasets from uk
  biobank,'' \emph{Neuroimage}, vol. 166, pp. 400--424, 2018.

\bibitem{miller2016multimodal}
K.~L. Miller, F.~Alfaro-Almagro, N.~K. Bangerter, D.~L. Thomas, E.~Yacoub,
  J.~Xu, A.~J. Bartsch, S.~Jbabdi, S.~N. Sotiropoulos, J.~L. Andersson
  \emph{et~al.}, ``Multimodal population brain imaging in the uk biobank
  prospective epidemiological study,'' \emph{Nature neuroscience}, vol.~19,
  no.~11, p. 1523, 2016.

\bibitem{zhang2001segmentation}
Y.~Zhang, M.~Brady, and S.~Smith, ``Segmentation of brain mr images through a
  hidden markov random field model and the expectation-maximization
  algorithm,'' \emph{IEEE transactions on medical imaging}, vol.~20, no.~1, pp.
  45--57, 2001.

\bibitem{smith2002fast}
S.~M. Smith, ``Fast robust automated brain extraction,'' \emph{Human brain
  mapping}, vol.~17, no.~3, pp. 143--155, 2002.

\bibitem{jenkinson2002improved}
M.~Jenkinson, P.~Bannister, M.~Brady, and S.~Smith, ``Improved optimization for
  the robust and accurate linear registration and motion correction of brain
  images,'' \emph{Neuroimage}, vol.~17, no.~2, pp. 825--841, 2002.

\bibitem{makropoulos2018developing}
A.~Makropoulos, E.~C. Robinson, A.~Schuh, R.~Wright, S.~Fitzgibbon, J.~Bozek,
  S.~J. Counsell, J.~Steinweg, K.~Vecchiato, J.~Passerat-Palmbach
  \emph{et~al.}, ``The developing human connectome project: A minimal
  processing pipeline for neonatal cortical surface reconstruction,''
  \emph{Neuroimage}, vol. 173, pp. 88--112, 2018.

\bibitem{fitzgibbon2020developing}
S.~P. Fitzgibbon, S.~J. Harrison, M.~Jenkinson, L.~Baxter, E.~C. Robinson,
  M.~Bastiani, J.~Bozek, V.~Karolis, L.~C. Grande, A.~N. Price \emph{et~al.},
  ``The developing human connectome project (dhcp) automated resting-state
  functional processing framework for newborn infants,'' \emph{NeuroImage},
  vol. 223, p. 117303, 2020.

\bibitem{schuh2018unbiased}
A.~Schuh, A.~Makropoulos, E.~C. Robinson, L.~Cordero-Grande, E.~Hughes,
  J.~Hutter, A.~N. Price, M.~Murgasova, R.~P.~A. Teixeira, N.~Tusor
  \emph{et~al.}, ``Unbiased construction of a temporally consistent
  morphological atlas of neonatal brain development,'' \emph{bioRxiv}, p.
  251512, 2018.

\bibitem{avants2011reproducible}
B.~B. Avants, N.~J. Tustison, G.~Song, P.~A. Cook, A.~Klein, and J.~C. Gee, ``A
  reproducible evaluation of ants similarity metric performance in brain image
  registration,'' \emph{Neuroimage}, vol.~54, no.~3, pp. 2033--2044, 2011.

\bibitem{tusor2017punctate}
N.~Tusor, M.~J. Benders, S.~J. Counsell, P.~Nongena, M.~A. Ederies,
  S.~Falconer, A.~Chew, N.~Gonzalez-Cinca, J.~V. Hajnal, S.~Gangadharan
  \emph{et~al.}, ``Punctate white matter lesions associated with altered brain
  development and adverse motor outcome in preterm infants,'' \emph{Scientific
  reports}, vol.~7, no.~1, pp. 1--9, 2017.

\bibitem{jack2008alzheimer}
C.~R. Jack~Jr, M.~A. Bernstein, N.~C. Fox, P.~Thompson, G.~Alexander,
  D.~Harvey, B.~Borowski, P.~J. Britson, J.~L.~Whitwell, C.~Ward \emph{et~al.},
  ``The alzheimer's disease neuroimaging initiative (adni): Mri methods,''
  \emph{Journal of Magnetic Resonance Imaging: An Official Journal of the
  International Society for Magnetic Resonance in Medicine}, vol.~27, no.~4,
  pp. 685--691, 2008.

\bibitem{tustison2010n4itk}
N.~J. Tustison, B.~B. Avants, P.~A. Cook, Y.~Zheng, A.~Egan, P.~A. Yushkevich,
  and J.~C. Gee, ``N4itk: improved n3 bias correction,'' \emph{IEEE
  transactions on medical imaging}, vol.~29, no.~6, pp. 1310--1320, 2010.

\bibitem{segonne2004hybrid}
F.~S{\'e}gonne, A.~M. Dale, E.~Busa, M.~Glessner, D.~Salat, H.~K. Hahn, and
  B.~Fischl, ``A hybrid approach to the skull stripping problem in mri,''
  \emph{Neuroimage}, vol.~22, no.~3, pp. 1060--1075, 2004.

\bibitem{modat2012inverse}
M.~Modat, M.~J. Cardoso, P.~Daga, D.~Cash, N.~C. Fox, and S.~Ourselin,
  ``Inverse-consistent symmetric free form deformation,'' in
  \emph{International Workshop on Biomedical Image Registration}.\hskip 1em
  plus 0.5em minus 0.4em\relax Springer, 2012, pp. 79--88.

\bibitem{peng2021accurate}
H.~Peng, W.~Gong, C.~F. Beckmann, A.~Vedaldi, and S.~M. Smith, ``Accurate brain
  age prediction with lightweight deep neural networks,'' \emph{Medical Image
  Analysis}, vol.~68, p. 101871, 2021.

\bibitem{jonsson2019brain}
B.~A. J{\'o}nsson, G.~Bjornsdottir, T.~Thorgeirsson, L.~M. Ellingsen, G.~B.
  Walters, D.~Gudbjartsson, H.~Stefansson, K.~Stefansson, and M.~Ulfarsson,
  ``Brain age prediction using deep learning uncovers associated sequence
  variants,'' \emph{Nature communications}, vol.~10, no.~1, pp. 1--10, 2019.

\bibitem{tsoi2015cognitive}
K.~K. Tsoi, J.~Y. Chan, H.~W. Hirai, S.~Y. Wong, and T.~C. Kwok, ``Cognitive
  tests to detect dementia: a systematic review and meta-analysis,'' \emph{JAMA
  internal medicine}, vol. 175, no.~9, pp. 1450--1458, 2015.

\bibitem{karras2020training}
T.~Karras, M.~Aittala, J.~Hellsten, S.~Laine, J.~Lehtinen, and T.~Aila,
  ``Training generative adversarial networks with limited data,'' \emph{arXiv
  preprint arXiv:2006.06676}, 2020.

\bibitem{sohoni2020no}
N.~S. Sohoni, J.~A. Dunnmon, G.~Angus, A.~Gu, and C.~R{\'e}, ``No subclass left
  behind: Fine-grained robustness in coarse-grained classification problems,''
  \emph{arXiv preprint arXiv:2011.12945}, 2020.

\bibitem{ba2016layer}
J.~L. Ba, J.~R. Kiros, and G.~E. Hinton, ``Layer normalization,'' \emph{arXiv
  preprint arXiv:1607.06450}, 2016.

\bibitem{captum2019github}
N.~Kokhlikyan, V.~Miglani, M.~Martin, E.~Wang, J.~Reynolds, A.~Melnikov,
  N.~Lunova, and O.~Reblitz-Richardson, ``Pytorch captum,''
  \url{https://github.com/pytorch/captum}, 2019.

\bibitem{Paszke2019PyTorch}
\BIBentryALTinterwordspacing
A.~Paszke, S.~Gross, F.~Massa, A.~Lerer, J.~Bradbury, G.~Chanan, T.~Killeen,
  Z.~Lin, N.~Gimelshein, L.~Antiga, A.~Desmaison, A.~Kopf, E.~Yang, Z.~DeVito,
  M.~Raison, A.~Tejani, S.~Chilamkurthy, B.~Steiner, L.~Fang, J.~Bai, and
  S.~Chintala, ``Pytorch: An imperative style, high-performance deep learning
  library,'' in \emph{Advances in Neural Information Processing Systems 32},
  H.~Wallach, H.~Larochelle, A.~Beygelzimer, F.~d~Alch{e}-Buc, E.~Fox, and
  R.~Garnett, Eds.\hskip 1em plus 0.5em minus 0.4em\relax Curran Associates,
  Inc., 2019, pp. 8024--8035. [Online]. Available:
  \url{http://papers.neurips.cc/paper/9015-pytorch-an-imperative-style-high-performance-deep-learning-library.pdf}
\BIBentrySTDinterwordspacing

\end{thebibliography}

\appendix
\counterwithin{figure}{section}
\section{Methods} 

\subsection{Network Architecture} 

\begin{figure*}[!ht]
  \centering
\makebox[\linewidth]{
	\includegraphics[width=1.0\textwidth]{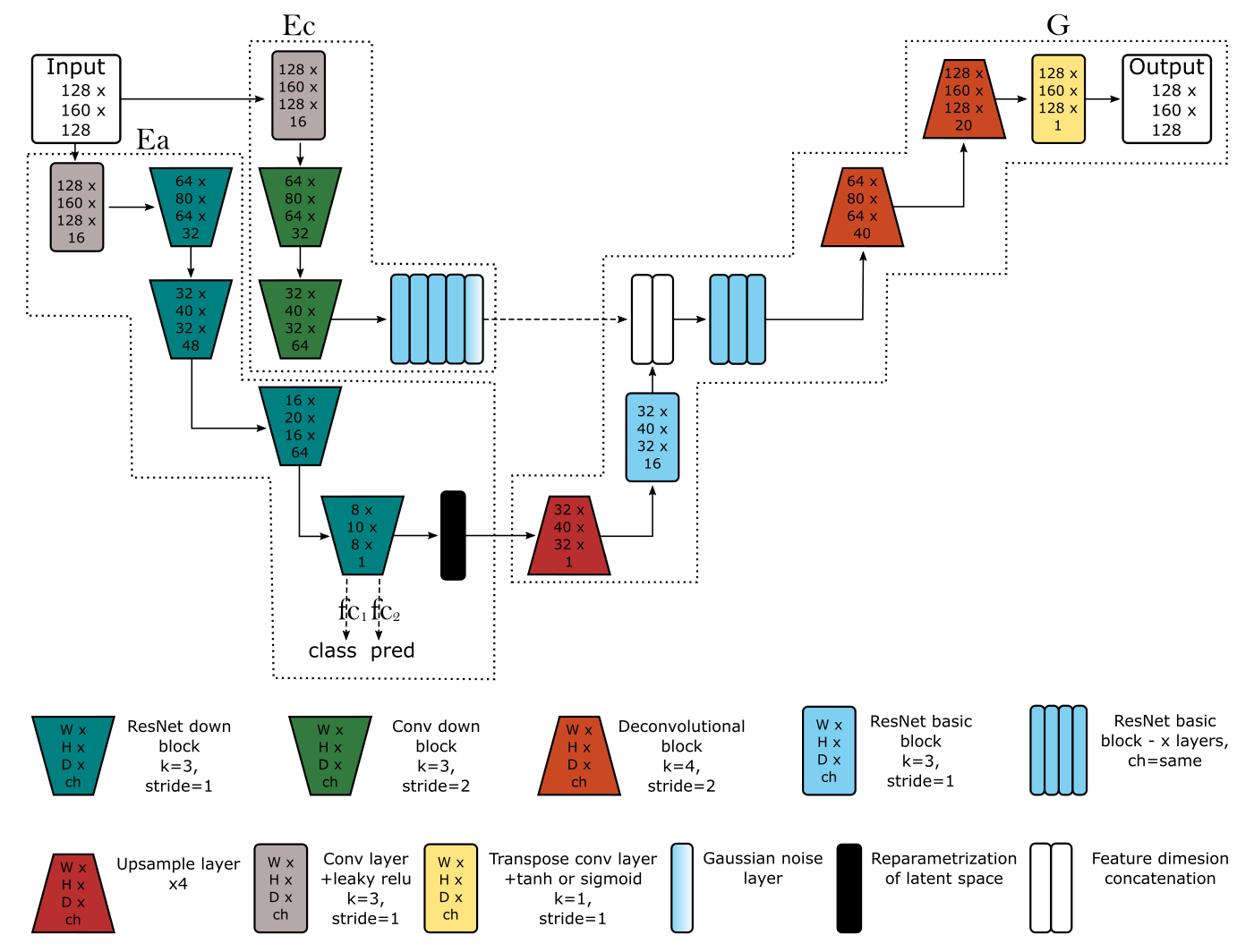}}
\caption{Network Architecture for 3D inputs.}
\label{fig:network_architecture}
\end{figure*}

Our encoder-decoder architecture for a 3D input is shown in Fig.~\ref{fig:network_architecture}. The architecture for a 2D input is the same, only using 2D convolutions and a 2D attribute space. Here, an input image is encoded using 2 shared networks, the attribute encoder $E_a$, and the content encoder $E_c$, and then is reconstructed or translated (to another class) using the generator, $G$. 

The key components of the attribute encoder include using down ResNet blocks (with average pooling, and leaky ReLU activation) for encoding the input image into a relatively large 3D latent space of size $8 \times 10 \times 8$ (in the 3D case), as opposed to a 1D vector, which is commonly seen in Variational Autoencoders (VAEs). We also added a fully connected layer ($f_{C_1}$) to the attribute latent space to enable classification. In our regression model $ICAM_{reg}$, we added an additional fully connected layer ($f_{C_2}$) to output a prediction. In early development, we found that using a 1D vector in the latent space was insufficient for encoding the required class information for brain imaging, and observed that some class information was instead encoded in the content encoder, which is meant to be class invariant. Using a sufficiently large 2D or 3D vector (depending on the input) helped with addressing this problem. 

The goal of the content encoder is to encode a class-irrelevant space, which allows translation between classes. The key components of the content encoder are using 2 down convolutional blocks (with instance normalisation, and ReLU activation), followed by 4 basic ResNet blocks (with instance normalisation, and ReLU activation), and finally a Gaussian noise layer. The basic ResNet blocks aids the encoding of a class-irrelevant space, and the Gaussian layer prevents the space from becoming zero.

Our generator takes in as input the content and attribute latent spaces. The attribute is first upsampled ($\times 4$, with nearest neighbors) to the same size as the content latent space, concatenated, and then combined using several basic ResNet blocks. Finally, we use deconvolutional blocks (transpose convolution with kernel size of 4, followed by average pooling, layer normalisation \cite{ba2016layer}, and a ReLU activation) to upsample to the original input size.

In addition, not shown in Fig.~\ref{fig:network_architecture}, our domain discriminator contains 6 convolutional layers with leaky ReLUs (kernel size = 3, stride = 2), followed by 2 additional convolutional layers (kernel size = 1, stride = 1), and adaptive average pooling for each class output, and real/ fake output. Our content discriminator contains 3 convolutional layers with leaky ReLUs (kernel size = 3, stride = 2), followed by an additional convolutional layer (kernel size = 4, stride = 1), adaptive average pooling, and a final fully connected layer for class output.

\subsection{Comparison methods - extra details}
We compare our proposed approach against a range of baselines in our experiments. For a fair comparison, we train and test all methods on the same training, validation and testing datasets.

\textbf{Grad-CAM, guided Grad-CAM \cite{selvaraju2017grad}, guided backpropagation (backprop) \cite{springenberg2014striving}, integrated gradients \cite{sundararajan2017axiomatic} and occlusion \cite{zeiler2014visualizing}: }
We trained a simple 3D ResNet with 4 down ResNet blocks, and a fully connected layer for classification. We then used the captum library \cite{captum2019github} implementation of Grad-CAM, guided Grad-CAM, guided backprop, integrated gradients and occlusion to generate the feature attribution maps for each method.  

Guided backprop \cite{springenberg2014striving} is a gradient-based method that computes the gradients with respect to an input image. More specifically it determines which pixels affect the prediction the most, by propagating only positive error signals (i.e. by applying ReLU to to the error during the backward pass).

Grad-CAM \cite{selvaraju2017grad} is gradient-based saliency method that computes the gradients of the target output with respect to the final convolutional layer of a network. The layer activations are weighted by the average gradient for each output channel and the results are summed over all channels to produce a coarse heatmap of prediction importance for each class. Guided Grad-CAM is simply the combination of the results of Grad-CAM and guided backprop.

Integrated gradients \cite{sundararajan2017axiomatic} is another method of analysing the gradient of the prediction output with respect to features of the input. It is defined as the integral of the gradients along the straight line path from a given baseline to the input image. A series of images are interpolated between the baseline (e.g. matrix of 0s) and the original image, and the integrated gradients are given by the integration of the computed gradients for all the images in the series.

Occlusion \cite{zeiler2014visualizing} is a perturbation-based method that involves replacing portions of an image with a block of a given baseline value (e.g. 0), and computing the difference in output. A heatmap is formed using the difference between the output probability attributed to the original volume and the probability computed for the occluded volume, for different positions of the occlusion block across the input image.

Grad-CAM was implemented on the last convolutional block of the ResNet, with a size of $4\times5\times4$, and was up-sampled to the input size for visualization. 
For the implementation of integrated gradients we considered a baseline volume with constant value of 0, and the integral was computed using 200 steps.
Occlusion was implemented using occlusion blocks with value 0, size $10\times10\times10$ and stride 5. 

\textbf{VA-GAN \cite{baumgartner2018visual}: }
We used the VA-GAN network for feature attribution, as described in the original paper.

\textbf{Model selection: } 
For VA-GAN and $ICAM$ the last model is selected in the Biobank experiments. For all our regression experiments the best model was selected based on the best MAE score on the validation dataset. In all other experiments, the models selected are based on the best model result on the validation dataset, using the NCC score. For Grad-CAM, guided Grad-CAM, guided backprop, integrated gradients and occlusion, as the FA maps are only generated after a network is trained, we could not select a model based on its performance with the NCC score, during training/ validation. We instead selected the best model based on the accuracy classification score on the validation dataset, to prevent the effect of overfitting. 

\subsection{Training details} 

We used PyTorch \cite{Paszke2019PyTorch} Python package in all of our deep learning experiments, and trained using NVIDIA TITAN GPUs. We trained our networks in a similar fashion to Lee et al. \cite{lee2019drit}. During training in each iteration, the content discriminator is updated twice, followed by the update of the encoders, generators, and domain discriminators (i.e. each training iteration uses 3 batches to perform these updates). For each update of the generator, an input is selected for each class (e.g. 2 inputs including class 0 and 1) to achieve translation. In addition, each input is encoded and translated to the opposite class by randomly sampling the attribute latent space, and obtaining an appropriate class, using the classifier. 

In all experiments, unless otherwise stated, we used the following hyperparameters during training of ICAM networks: learning rate for content discriminator = 0.00004, learning rate for the rest = 0.0001, Adam optimiser with betas = (0.5, 0.999), $\lambda_{D^c}=1, \lambda_{D}=1, \lambda_{BCE}=10, \lambda_{KL}=0.01, \lambda_{M}=10, \lambda_{z^a}=1, \lambda_{rec}=100, \lambda_{D_{BCE}}=1$ for discriminator optimisation, and $\lambda_{D_{BCE}}=5$ for generator optimisation. We do not use augmentation techniques in any of our experiments.

In the UK Biobank experiments, we trained all networks for 50 epochs. In the ADNI experiments, all networks (including VA-GAN) were trained for 300 epochs. In the ADNI experiments, because we had a limited dataset, we further refined ICAM with updated lambdas ($\lambda_{rec}=10$, and $\lambda_{BCE}=20$) for another 200 epochs. We could not refine VA-GAN any further because generator and discriminator losses went to zero during training, often after 150 epochs. In our dHCP experiments, we used the same hyperparameters as in our Biobank experiments, but trained the networks for 1000 epochs. 

\textbf{Regression models: }
For training our regression models (including the baseline CNN), we used a pretrained network using the classification model (training as described above). All networks were then retrained with the same hyperparamaters as before, with the addition of the regression loss. In our dHCP experiments we used a model first pretrained on the biobank dataset.

\textbf{Baseline methods: } For training VA-GAN, and DRIT, we used the default parameters as provided in the original papers and publicly released code repositories. For Grad-CAM, integrated gradients, and occlusion, the classifier network was trained with learning rate of 0.0001, SGD with momentum of 0.9, for 50 epochs, and using a weighted BCE loss to account for class-unbalanced training data. Since the model converged by 50 epochs, we did not train for any longer. 

For training the baseline-CNN for the dHCP experiments, we used the same architecture as our attribute encoder, and trained using smooth L1 loss, with Adam optimiser (learning rate = 0.0001, betas = [0.5, 0.999]) for 1000 epochs. 

\end{document}